\documentclass[10pt,twocolumn]{article}

\usepackage[utf8]{inputenc}
\usepackage[T1]{fontenc}
\usepackage{lmodern}
\usepackage[margin=1in]{geometry}
\usepackage{amsmath, amssymb, amsthm, mathtools}
\usepackage{graphicx}
\usepackage{subcaption}
\usepackage{booktabs}
\usepackage{multirow}
\usepackage{adjustbox} % auto-scale wide tables
\usepackage{xcolor}
\usepackage{enumitem}
\usepackage{siunitx}
\usepackage{hyperref}
\usepackage[numbers,sort&compress]{natbib}
\usepackage{authblk}
\usepackage{listings}
\usepackage{balance}    % balance the last page columns
\usepackage{dblfloatfix}% improve two-column floats

\newcommand{\CB}{\mathbf{C}}        % Big-brain context tokens
\newcommand{\Vt}{\mathbf{V}}        % Vision tokens
\newcommand{\Wt}{\mathbf{W}}        % Text tokens
\newcommand{\st}{\mathbf{s}}        % State token
\newcommand{\qk}{\mathbf{q}}        % Action query
\newcommand{\Hmax}{\mathcal{H}_{\max}} % entropy threshold
\newcommand{\Nc}{N_{\!c}}           % number of context tokens

\newcommand{\dmodel}{d}             % unified model dimension
\newcommand{\Dact}{D}               % action dimension
\newcommand{\Kchunk}{K}             % parallel steps

\newcommand{\SRcn}{\mathrm{SR}_{\mathrm{cn}}} % compute-normalized success rate
\newcommand{\Cbudget}{\mathcal{C}}           % compute budget per time unit

\newif\ifshowresults
\showresultstrue

\newcommand{\headname}{ParaCAT} % Parallel Categorical Action Transformer
\newcommand{\Pons}{Pons Adapter}

\title{SaiVLA-0: Cerebrum--Pons--Cerebellum Tripartite Architecture for Compute-Aware Vision-Language-Action}

\author[1]{Xiang Shi}
\author[1]{Wenlong Huang}
\author[1]{Menglin Zou}
\author[1]{Xinhai Sun}
\affil[1]{Synthoid.ai}
\affil[ ]{\texttt{contact: scholar@synthoid.ai}}

\date{March 2026}

\begin{document}

\maketitle

% Abstract (Concept paper; placeholders allowed)
\begin{abstract}
We revisit Vision‑Language‑Action through a neuroscience‑inspired triad. Biologically, the \emph{Cerebrum} provides stable high‑level multimodal priors and remains frozen; the \emph{Pons Adapter} integrates these cortical features with real‑time proprioceptive inputs and compiles intent into execution‑ready tokens; and the \emph{Cerebellum} (ParaCAT) performs fast, parallel categorical decoding for online control, with hysteresis/EMA/temperature/entropy for stability. A fixed‑ratio schedule and two‑stage feature caching make the system compute‑aware and reproducible. Inspired by active, foveated vision, our wrist ROIs are geometrically tied to the end‑effector via calibrated projection, providing a movement‑stabilized, high‑resolution view that is sensitive to fine‑grained pose changes and complements the global context of the main view. The design is modular: upgrading the Cerebrum only retrains the Pons; changing robots only trains the Cerebellum; cerebellum‑only RL can further refine control without touching high‑level semantics. As a concept-and-protocol paper with preliminary evidence, we outline a timing protocol under matched conditions (GPU/resolution/batch) to verify anticipated efficiency gains. We also report preliminary LIBERO evidence showing that split feature caching reduces training time (7.5h\(\to\)4.5h) and improves average success (86.5\%\(\to\)92.5\%) under official N1.5 head-only training, and that SaiVLA0 reaches 99.0\% mean success.
\end{abstract}

% Keywords (optional for arXiv)

\section{Introduction}
Modern VLA models often entangle semantic understanding and high-frequency control in a single system, leading to high latency and instability, especially under limited-data regimes where end-to-end fine-tuning of large VLMs is impractical and risks overfitting \cite{kim2024openvla,black2024pi0,liu2025rdt1b,chi2023diffusionpolicy}~\cite{ahn2022saycan,driess2023palme}. Relying solely on last-layer representations also struggles to simultaneously capture global semantics and local geometric and contact details, and inconsistent prompts/calibration impede reproducibility.

We revisit VLA through a neuroscience‑inspired triad that separates understanding from fast control while keeping compute usage explicit and controllable \cite{barto2003hrl,botvinick2009hierarchical,debenedictis2022cerebellum}. The \emph{Cerebrum} provides stable, high‑level multimodal priors and stays frozen during downstream learning. The \emph{\Pons{}} mirrors the pons by integrating cortical representations with real‑time perceptual and proprioceptive inputs, compiling intent into execution‑ready tokens. The \emph{Cerebellum} (ParaCAT) then performs fast, parallel categorical decoding to adjust action policies online under tight latency. This sensorimotor analogy also motivates our two‑stage training (Stage~A caches frozen Cerebrum features; Stage~B trains the Pons–Cerebellum pathway end‑to‑end). Similar dual‑system ideas have appeared in industrial humanoid stacks such as Figure AI's Helix~\cite{figure2024helix}.

\begin{figure*}[t]
  \centering
  \includegraphics[width=\linewidth]{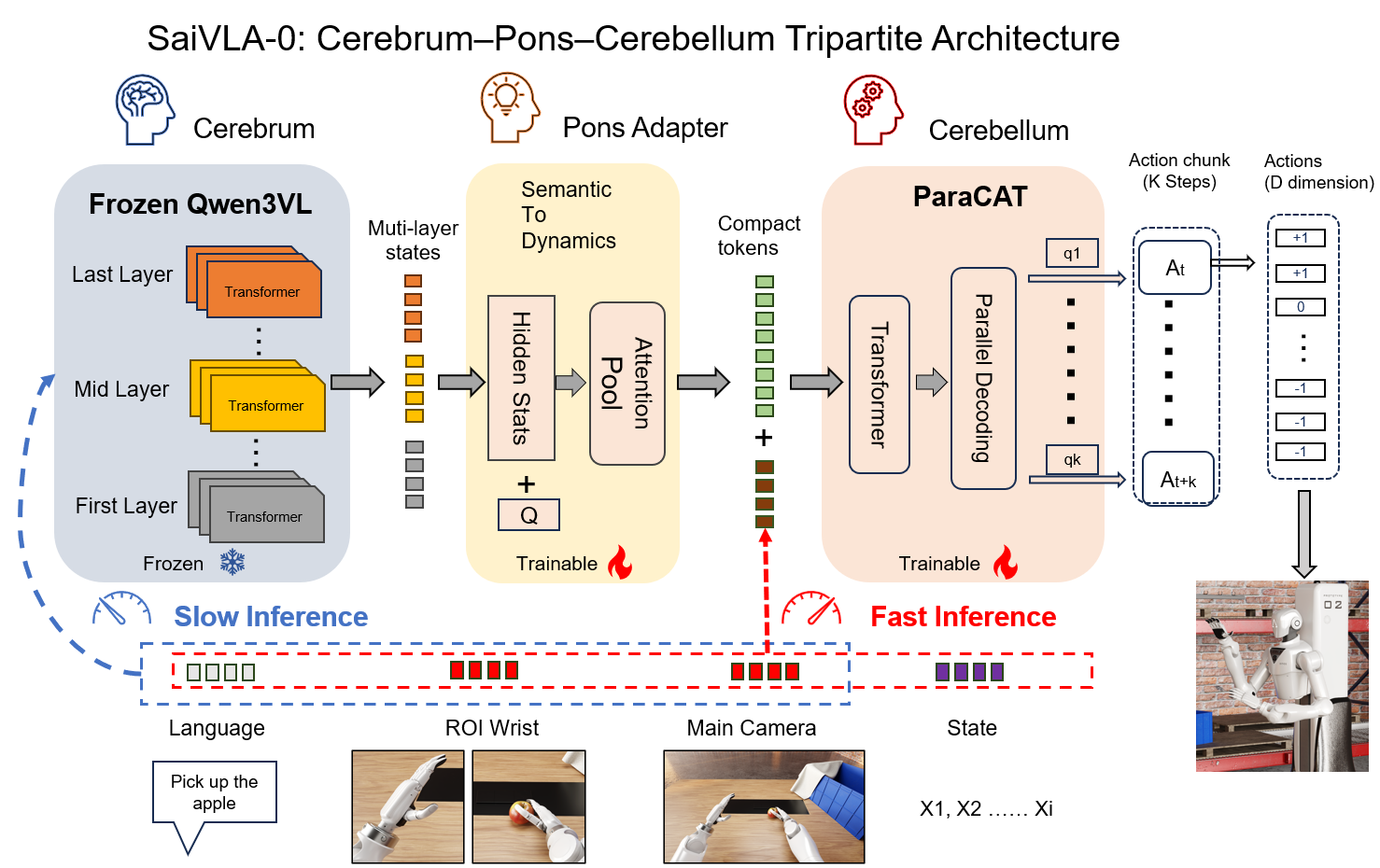}
  \caption{Overview: Frozen Cerebrum emits context tokens sparsely; the Cerebellum fuses image+text+state+brain tokens at high frequency and outputs categorical deltas.}
  \label{fig:intro-overview}
\end{figure*}

Concretely, a frozen large VLM (Cerebrum) runs at low frequency and exposes multi‑layer hidden states. The \Pons{} projects them into a small set of context tokens. The Cerebellum---a ViT + text encoder + \headname{}---runs at high frequency, fusing (i) the current image (main \(1028{\times}800\!\to\!256^2\) with two wrist ROIs \(256^2\)), (ii) the instruction, (iii) robot state, and (iv) Cerebrum tokens to produce per‑dimension categorical deltas \(\{-1,0,+1\}\). We adopt a \textbf{fixed‑ratio schedule} (Cerebrum every \(N{=}5\) chunks) with \textbf{micro‑horizon reuse} (\(K{=}20\) steps/forward), plus hysteresis/EMA/temperature/entropy for stability under latency. An overview is shown in Figure~\ref{fig:intro-overview}.

We adopt feature caching and a two-stage training pipeline: (A) offline Cerebrum inference and caching of multi-layer tokens + prompt meta; (B) training the Cerebellum and the Pons Adapter on cached features and current frames; optionally (C)(optional) lightly tuning adapters. This yields faster iteration and better reproducibility \cite{nvidia2025gr00tn1,nvidia2025gr00tn1_5,kim2024openvla}.

Neuroscience-inspired ROI. Human vision is foveated: the fovea is continuously directed toward task‑relevant targets and provides high‑acuity detail, while peripheral vision supplies global context. Our ROI design mirrors this: wrist ROIs are geometrically bound to the end‑effector via calibrated projection (akin to retinotopic mapping), offering a movement‑stabilized, high‑resolution view that captures fine‑grained pose and contact changes. This perspective is complementary to recent work on visual foundation models for embodied AI, which study “artificial visual cortex” representations reusable across tasks and embodiments~\cite{majumdar2023vc}. ROI tokens are fused with main‑view tokens via cross‑attention, resembling attentional gating across foveal/peripheral pathways; when ROI confidence drops (e.g., occlusion), we fall back to the main view and adopt a more conservative decoding policy (higher temperature/stronger hysteresis), analogous to risk‑aware visuomotor behavior under uncertainty.

The split is \emph{compute‑aware}: latency is reported by component (Cerebrum once‑call vs Cerebellum per‑forward), we expose throughput knobs \(N\) and \(K\), and we standardize compute‑normalized success \(\SRcn\). By “compute‑aware”, we mean that latency, FLOPs, and success are always reported jointly, including \(\SRcn\) for fair comparison. The modularity further enables cerebellum‑only RL in simulation without touching the Cerebrum or the \Pons{}.

\textbf{Contributions.}
1) Foveated, geometry‑tied ROI: end‑effector poses are projected into the main view via calibration, yielding wrist ROIs that remain stable in the tool frame and capture fine‑grained pose and contact changes; ROI is fused with the main view and falls back gracefully under low confidence.\\
2) Precision‑control tasks: we introduce quantitative goals (e.g., "move the object left by 10 cm") to enforce measurable spatial understanding and fine‑grained control.\\
3) Efficient separated training: Stage‑A offline Cerebrum caching + Stage‑B Cerebellum and adapter training; we report preliminary training-time gains on LIBERO and provide a public timing protocol for verification under matched conditions.\\
4) \headname{} (Parallel Categorical Action Transformer) head: parallel softmax categorical decoding produces \(K\) steps in one forward; we provide a timing protocol under matched GPU/resolution/batch to verify anticipated efficiency gains, and leave full coverage to future work.\\
5) Modular upgradability and transfer: upgrading the Cerebrum only requires retraining a lightweight adapter; changing robots while reusing the same Cerebrum only requires training the Cerebellum, improving generality and maintainability.\\
We further adopt a fixed‑ratio schedule (\(N{=}5\)) and micro‑horizon reuse (\(K{=}20\)), and report compute‑normalized metrics (\(\SRcn\)) with explicit latency breakdowns.

\textbf{Evaluation plan and concept positioning.} We outline evaluation on LIBERO‑Spatial/Object/Goal/Long (10 tasks each, 500 episodes per subset) and three real/precision suites (folding clothes: 5 tasks/200 episodes; put X into pot: 10/400; move by fixed distance: 10/200). We will report success, jitter/jerk, \(f_{\text{fwd}}\), \(f_{\text{eff}}\), \(\SRcn\), and latency splits. This is a concept-and-protocol paper with preliminary evidence: we articulate design hypotheses, report stage-wise results (Section~5), and publish a reproducible evaluation protocol (evaluation scripts, cache schema, success criteria) to facilitate independent verification, rather than claiming conclusive superiority. We include stage-wise LIBERO evidence (Section~5): split feature caching improves success and reduces training time under official N1.5 head-only training, backbone swapping (Eagle2.5 vs Qwen3VL-2B) shows consistent trends under controlled settings, and SaiVLA0 reaches 99.0\% mean success on LIBERO. LIBERO is used as a phase-wise validation; the core target remains real-robot SaiVLA0 with more tasks and larger data, following the original design in this paper.

% Figure 1: Overview (external image)

\section{Related Work}
We situate our work among the following strands.

\paragraph{VLA and Robot Foundation Models.}
Representative systems include GR00T variants (e.g., GR00T‑N~1.5), OpenVLA, OpenVLA‑OFT, $\pi^0$, and RDT‑1B~\cite{nvidia2025gr00tn1,nvidia2025gr00tn1_5,kim2024openvla,kim2025oft,black2024pi0,liu2025rdt1b}, as well as language‑conditioned manipulation systems such as CLIPort and MOO‑style open‑world object manipulation policies~\cite{shridhar2022cliport,stone2023openworld}. These efforts scale data and model sizes and often favor end‑to‑end finetuning of large backbones for generalization. While effective, such pipelines can be compute‑intensive and brittle in limited‑data regimes, and they conflate high‑level understanding with low‑latency control within a single stack.
These systems relate to web‑scale generalist robot policies such as RT‑1/RT‑2 and Open X‑Embodiment~\cite{brohan2022rt1,brohan2023rt2,openxembodiment2023}.
Unlike end-to-end stacks, we make the \emph{pons} explicit as a learnable compiler that reparameterizes the action–semantic manifold before cerebellar execution.

\paragraph{Freezing, Adapters and Representation Alignment.}
Freezing backbones with lightweight adapters (LoRA/Adapters) and representation alignment (e.g., CCA/Procrustes)~\cite{hu2022lora,houlsby2019adapter,hotelling1936cca,gower1975procrustes} balances quality and efficiency. Recent work on language‑ and video‑driven pretraining further supports this frozen‑backbone paradigm, e.g., R3M and Voltron, which learn reusable visual representations for robot manipulation from human videos and language~\cite{nair2023r3m,karamcheti2023voltron}. We go further by \emph{fully freezing} the Cerebrum (frozen VLM) and training only a Pons Adapter that fuses multi‑layer features into a compact context‑token set \(\CB\). To ensure reproducibility and iterative speed, we adopt a two‑stage pipeline with versioned feature caches. \emph{Unlike prior freezing works that still fine‑tune large heads, our Pons\,+\,Cerebellum are lightweight, enabling two‑stage caching and compute‑normalized reporting aligned with our contributions.}

\paragraph{Action Representations and Decoders.}
Continuous regression, diffusion/flowmatching heads, and parallel/categorical decoders (e.g., OFT)~\cite{chi2023diffusionpolicy,black2024pi0,liu2025rdt1b,kim2025oft,zhao2023act,pertsch2025fast} offer different trade‑offs among latency, stability, and calibration. We adopt \emph{\headname{}}, a Parallel Categorical Action Transformer head that outputs per‑dimension \(\{-1,0,+1\}\) deltas with softmax decoding and stabilizes execution via hysteresis/EMA/temperature/entropy controls. This favors low latency and calibrated switching, while we acknowledge an ultimate precision ceiling (addressed as future hybrid heads). \emph{Our ParaCAT performs parallel softmax decoding to produce K steps in one forward, directly supporting low‑latency, compute‑aware control.}

\paragraph{Multi-layer Feature Fusion.}
Layer‑wise extraction and cross‑layer attention supply complementary semantics and geometry. We select early/mid/late Cerebrum layers, fuse and compress them into a small set of context tokens \(\CB\) consumed by the Cerebellum.

\paragraph{Hierarchical \& Tripartite Control.}
Hierarchical planning plus low‑level control~\cite{barto2003hrl,botvinick2009hierarchical} motivates separating low‑frequency semantics from high‑frequency actuation. Our tripartite architecture makes this separation \emph{compute‑aware}: a fixed‑ratio schedule (Cerebrum every \(N{=}5\) chunks) with micro‑horizon reuse (\(\Kchunk{=}20\) steps/forward) exposes explicit throughput–reactivity trade‑offs, reported via latency splits and compute‑normalized success \(\SRcn\). Similar separation between cortical planning and fast cerebellar execution has been studied in neuroscience~\cite{debenedictis2022cerebellum,vanes2019cerebellum} and appears in industrial humanoid controllers such as Figure AI's Helix~\cite{figure2024helix}.

\paragraph{Structured Prompts \& ROI.}
Structured/JSON prompts stabilize Cerebrum outputs; ROI/multi‑view perception enhances local contact cues. We use JSON prompts (fields: goal/constraints/objects/failure\_cases/environment; 50\% field shuffle during training) and dual wrist ROIs (\(256^2\) each) fused by the Cerebellum. \emph{Distinct from fixed wrist cameras, our geometric ROI is tied to the tool frame via calibrated projection, improving pose‑change sensitivity and aligning directly with our contributions.}

\begin{table}[t]
  \centering
  \caption{Comparison axes (high-level): data demand, latency, reproducibility cost, hierarchical/tripartite, async scheduling support. Baselines are instantiated using GR00T~\cite{nvidia2025gr00tn1,nvidia2025gr00tn1_5}, OpenVLA~\cite{kim2024openvla}, OpenVLA-OFT~\cite{kim2025oft}, and diffusion-style policies~\cite{chi2023diffusionpolicy,black2024pi0,liu2025rdt1b}.}
  \label{tab:rw-compare}
  \begin{adjustbox}{max width=\linewidth}
  \begin{tabular}{lccccc}
    \toprule
    Method & Data & Latency & Reprod. Cost & Hier./Dual & Async \\
    \midrule
    GR00T-like~\cite{nvidia2025gr00tn1} & High & High & High & \(\times\) & \(\times\) \\
    OpenVLA~\cite{kim2024openvla} & High & Med & Med & \(\times\) & \(\times\) \\
    OFT~\cite{kim2025oft} & Med & Med & Med & \(\times\) & \(\times\) \\
    Diffusion~\cite{chi2023diffusionpolicy} & High & High & High & \(\times\) & \(\times\) \\
    \textbf{SaiVLA-0 (ours; target)} & Low (target) & Low (target) & Low (target) & \(\checkmark\) & \(\checkmark\) \\
    \bottomrule
  \end{tabular}
  \end{adjustbox}
\end{table}

\paragraph{Indirect evidence from prior art.}
Findings in hierarchical control, freezing/adapters, multi‑layer fusion, and categorical/discrete control collectively suggest gains in stability and latency under constrained data/compute. Our work consolidates these signals into a unified tripartite architecture with fixed‑ratio scheduling and two‑stage caching; Sections~3–5 specify the architecture and evaluation protocol under matched training steps and consistent evaluation settings.

\section{Method}
We describe the tripartite architecture and its components. Notation aligns with the overview in the main text; we restate for completeness.

\paragraph{Overall Architecture (Tripartite).}
The frozen VLM (Cerebrum) runs sparsely and exposes multi-layer hidden states. A trainable Pons Adapter fuses them into fixed-length context tokens \(\CB \in \mathbb{R}^{\Nc \times \dmodel}\). The high-frequency Cerebellum fuses image tokens \(\Vt\), text tokens \(\Wt\), and a state token \(\st\); its action head introduces \(\Kchunk\times \Dact\) learnable action queries \(\{\qk_{k,j}\}_{k=1..\Kchunk,\,j=1..\Dact}\). These queries are model parameters (learnable embeddings), not exogenous inputs. The Transformer processes the following internal token sequence
\[
\mathbf{X} = [\CB; \Vt; \Wt; \st; \{\qk_{k,j}\}_{k=1..\Kchunk,\,j=1..\Dact}].
\]
The cerebellum produces per-dimension categorical logits \(\mathbb{R}^{3}\) (for \(-1/0/+1\)) and uses hysteresis/EMA/temperature/entropy constraints for stability under latency. In our dual-arm hardware, \(\Dact{=}16\) (two 7-DoF arms plus two gripper open/close DoFs) \cite{barto2003hrl,debenedictis2022cerebellum,figure2024helix}.

\paragraph{Design hypotheses (testable predictions).}
H1 (Tripartite latency--stability): Separating a low-frequency frozen Cerebrum from a high-frequency Cerebellum reduces jitter and end-to-end latency at comparable success under limited-data regimes.\\
H2 (Multi-layer context): Using early/mid/late Cerebrum tokens improves contact-sensitive behaviors over last-layer-only context.\\
H3 (Categorical control): Per-dimension \(\{-1,0,+1\}\) deltas with hysteresis/EMA/temperature/entropy improve calibration and reduce oscillation versus continuous heads at similar compute.\\
H4 (Two-stage caching): Offline Cerebrum caching reduces wall-clock and seed variance versus end-to-end tuning at similar accuracy.\\
H5 (Scheduling): A simple fixed Cerebrum cadence (one call per \(N\) cerebellum chunks) amortizes compute while preserving success under the same budget.\\
H6 (Foveated ROI): Geometry-tied wrist ROIs plus the main view improve contact-sensitive behavior and stability metrics (e.g., jitter and jerk) over main-view-only perception at comparable compute.\\
H7 (Compute-normalized reporting): Compute-normalized success \(\SRcn\), together with latency and effective action-rate metrics, yields more meaningful comparisons across heads and schedules than raw success alone under fixed wall-clock budgets.

\paragraph{Cerebrum (Frozen VLM).}
\emph{Backbone and Prompt.} We freeze a large VLM (Qwen-VL-8B in our main setup; 4B/32B are used in scaling studies), including the tokenizer and prompt template. We use a structured/JSON prompt (fields: goal, constraints, objects, failure cases, environment) and randomize field order during training to improve robustness.

\emph{Multi-layer Outputs (Frozen).}
We expose early/mid/late hidden states \(\{H_B^{(l)}\}_{l\in\{l_1,l_m,l_L\}}\) from the frozen VLM and keep all backbone/tokenizer/prompt parameters fixed. 
Projection/fusion/pooling are implemented by a trainable \textbf{Pons Adapter (Brain-to-Cerebellum Adapter)} on the cerebellum side (see below), so that the Cerebrum module remains entirely frozen.
Design note: early/mid/late layers capture edges/shapes, object/part clues, and semantics/tasks respectively.

\paragraph{Pons Adapter.}
\emph{\Pons{} (semantic-to-dynamics compiler; trainable, online).}
Given multi-layer outputs \(\{H_B^{(l)}\}\), the \Pons{} projects, fuses, and summarizes them into \(\CB\in\mathbb{R}^{\Nc\times\dmodel}\) that serve as context tokens to the action head. Concretely, the \Pons{} (i) performs high-dimensional sparse recoding of structured Cerebrum intent into execution-ready tokens, (ii) factorizes action structure into geometry, dynamics priors, and control objectives to form composable motor primitives, and (iii) aligns feedback and intent to ease cerebellar forward-model updates for closed-loop stability. We use layer-wise projections and fusion (GLU and cross-layer attention), followed by \textbf{attention token pooling} with learnable queries:
\[
\begin{aligned}
Q &\in \mathbb{R}^{\Nc\times \dmodel},\\
K_a &= G W_k,\\
V_a &= G W_v,\\
A &= \mathrm{softmax}\!\left(\frac{QK_a^\top}{\sqrt{\dmodel}}\right) \in \mathbb{R}^{\Nc\times T},\\
\CB &= \mathrm{LN}(A V_a).
\end{aligned}
\]
The \Pons{} is trained \emph{jointly} with the cerebellum (online adapter); only the Cerebrum is frozen. In our two-stage setup, Stage A caches the frozen Cerebrum multi-layer outputs \(\{H_B^{(l)}\}\), and Stage B trains the \Pons{} and the action head end-to-end on cached Cerebrum features and current frames.
\emph{Inputs.} A ViT encodes the current RGB main view (captured at \(1028{\times}800\) and resized to \(256{\times}256\)) and two wrist ROIs (each \(256{\times}256\)) into \(\Vt\); a (frozen or lightly adapted) text encoder maps the instruction to \(\Wt\); low-dimensional robot state becomes \(\st\). We tag modalities (brain/image/text/state/action) via learnable embeddings and adopt relative positional encodings for variable token lengths.

\paragraph{Cerebellum. \headname{} (Parallel Categorical Action Transformer; \(\Kchunk\times\Dact\) queries).}
An encoder-only Transformer processes \(\mathbf{X}\). \headname{} performs parallel softmax categorical decoding: for each time--dimension query \((k,j)\), we extract the final hidden \(\mathbf{z}_{k,j}\in\mathbb{R}^{\dmodel}\) and map it to categorical logits via a shared lightweight head:
\[
\begin{aligned}
\mathbf{o}_{k,j} &= W_{\text{out}} \mathbf{z}_{k,j} + \mathbf{b}_{\text{out}} \in \mathbb{R}^{3},\\
\mathbf{p}_{k,j} &= \mathrm{softmax}(\mathbf{o}_{k,j}/\tau),\\
&\text{for } k=1..\Kchunk,\ j=1..\Dact.
\end{aligned}
\]
Choosing a ternary grid \(\{-1,0,+1\}\) keeps the label space extremely simple, which empirically stabilizes optimization and makes the head easy to train under limited data and tight latency budgets. It also matches the discriminative nature of the frozen VLM backbone: the Cerebellum only decides whether each control dimension should move negatively, stay, or move positively, while the actual metric scale is set by a fixed step size \(\boldsymbol{\delta}\). From a neuroscience perspective, these categorical deltas are analogous to discrete spike events: after temporal integration and population-like aggregation via EMA and hysteresis, they give rise to smooth, continuous control signals at the joint and muscle level.
We train with class-weighted cross-entropy, label smoothing, and optional temporal smoothness:
\[
\begin{aligned}
\mathcal{L} &= \sum_{k=1}^{\Kchunk}\sum_{j=1}^{\Dact} w_j \cdot \mathrm{CE}\big(\mathbf{p}_{k,j}, \mathrm{LabelSmooth}(y_{k,j},\epsilon)\big)\\
&\quad + \lambda_H \sum_{k,j}\mathcal{H}(\mathbf{p}_{k,j})
 + \lambda_T \sum_{k=2}^{\Kchunk}\sum_{j}\mathrm{KL}\big(\mathbf{p}_{k,j}\,\|\,\mathrm{sg}(\mathbf{p}_{k-1,j})\big).
\end{aligned}
\]
\noindent For execution, hysteresis thresholds and EMA reduce jitter:
\[
\hat{\Delta}_{t,j}=
\begin{cases}
 +1 & \text{if } p_{t,j}^{(+1)}-p_{t,j}^{(0)}>\theta_{\uparrow} \\
 -1 & \text{if } p_{t,j}^{(-1)}-p_{t,j}^{(0)}>\theta_{\downarrow} \\
 \hat{\Delta}_{t-1,j} & \text{otherwise}
\end{cases}
\]
\[
u_t = \alpha u_{t-1} + (1-\alpha)\big(\hat{\Delta}_{t}\odot \boldsymbol{\delta}\big).
\]
\textit{Execution policy (default: micro-horizon reuse without triggers).}
Given the \((\Kchunk\times \Dact)\) predictions from one forward pass, we execute steps sequentially \(k{=}1,2,\dots,\Kchunk\) without re-forwarding, and then perform a new forward pass for the next chunk. This \emph{execute-and-reuse} policy amortizes inference and yields a higher effective action rate while keeping implementation simple; we do not use uncertainty/deviation-based early re-planning in this work. In practice, \headname{} delivers single-forward multi-step decisions and enables substantial inference speedups versus diffusion/flowmatching heads \cite{kim2025oft,zhao2023act,chi2023diffusionpolicy,black2024pi0,liu2025rdt1b,pertsch2025fast}.

\paragraph{Dual-Frequency Scheduling (fixed ratio).}
We use a simple fixed-rate schedule: the Cerebrum is invoked once every \(N\) cerebellum chunks (default \(N{=}5\)). Let \(f_{\text{fwd}}\) be the cerebellum forward rate and \(\Kchunk\) the chunk size; the effective action rate is \(f_{\text{eff}}\!\approx\!\Kchunk\cdot f_{\text{fwd}}\), while the Cerebrum amortizes over \(N\) chunks.
This policy is simple and effective in practice; we omit pseudocode for brevity.

\paragraph{Feature Caching and Two-Stage Training.}
\textbf{Stage A (Frozen Cerebrum caching).} Run the frozen VLM offline to extract and store multi-layer outputs \(\{H_B^{(l)}\}\) together with prompt meta, calibration, and trajectories in \texttt{npz/mmap} (with version hashes, shapes, timestamps, checksums, dependencies). 
\textbf{Stage B (Online adapter + cerebellum training).} Train the Pons Adapter and the ViT+Transformer action head jointly on cached \(\{H_B^{(l)}\}\) and current frames; the Cerebrum remains frozen. 
\textbf{Stage C (optional).} Lightly tune small bridging layers if needed. We provide a cache validation checklist (Appendix). Viewed through the sensorimotor lens, this split corresponds to caching cortical outputs (Stage~A) and adapting the pons–cerebellum pathway online (Stage~B) \cite{hu2022lora,houlsby2019adapter,kim2024openvla,nvidia2025gr00tn1}.

\begin{figure}[t]
  \centering
  \includegraphics[width=\linewidth]{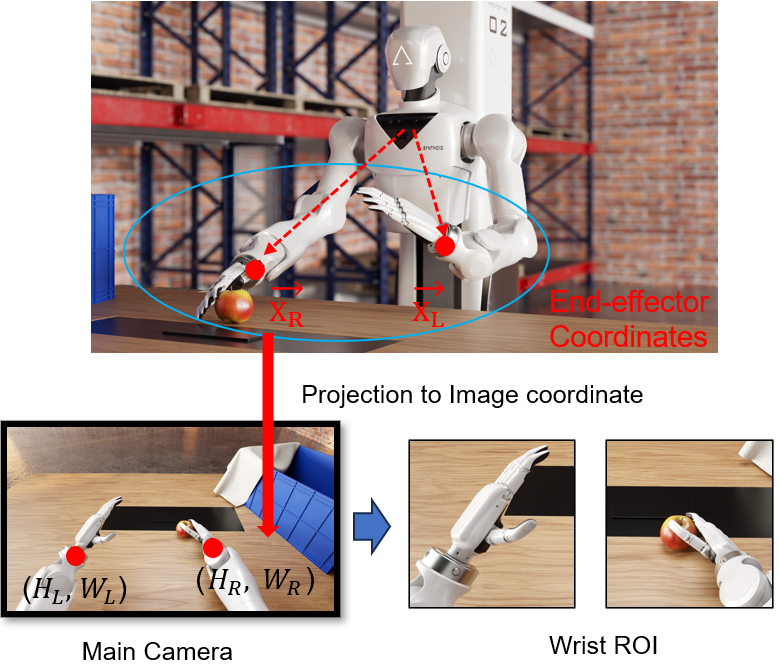}
  \caption{ROI projection pipeline. End-effector poses are projected into image coordinates via calibrated intrinsics/extrinsics to produce geometry-tied wrist crops; ROIs complement the main view with movement-stabilized, high-resolution contact cues.}
  \label{fig:roi-proj}
\end{figure}

\paragraph{ROI / Multi-View Integration.}
We project end-effectors into image coordinates via calibrated intrinsics and crop \textbf{two wrist ROIs} (left/right). Unlike a fixed wrist camera, these ROIs are \emph{geometrically tied to the end-effector}, remaining stable relative to the tool frame and thus more sensitive to small pose and contact changes beyond mere distance cues. We fuse ROI and main-view tokens via cross-attention; under low ROI confidence/occlusion, we fall back to the main view and adopt a more conservative decoding policy (higher temperature and stronger hysteresis/EMA). An illustration of the projection pipeline is shown in Figure~\ref{fig:roi-proj}.

\noindent\textit{Neuroscience inspiration.}
Human vision is foveated: the fovea is continuously directed toward task-relevant targets (hands or tools) and provides high-acuity detail, while peripheral vision supplies coarse global context. Our ROI functions as a foveal view that is dynamically aligned to the end-effector (via retinotopic-like projection), whereas the main view acts as peripheral context. Cross-attention mirrors attentional gating across visual pathways; confidence-aware fallback resembles risk-aware visuomotor behavior under uncertainty~\cite{debenedictis2022cerebellum,vanes2019cerebellum}.

\paragraph{Complexity and Latency.}
We separate Cerebrum once-call cost and Cerebellum per-step cost. Let \(f\) be the control loop frequency and \(N\) the Cerebrum recurrence interval. We define a compute budget per time unit
\[
\Cbudget = \frac{1}{N}\mathrm{FLOPs}_{\text{brain\_once}} + f \cdot \mathrm{FLOPs}_{\text{cere\_per\_step}},
\]
and report compute-normalized success \(\SRcn = \mathrm{SuccessRate}/\Cbudget\). We also report cold Cerebrum latency, per-step Cerebellum latency (incl.\ ROI), and achieved closed-loop frequency. Default hyperparameters are summarized in Appendix (Main defaults) and Table~\ref{tab:dt-hypers}.

\paragraph{Amortized compute with chunk reuse.}
Under the execute-and-reuse policy (no early re-planning), one Cerebellum forward covers \(\Kchunk\) control steps. If \(f_{\text{fwd}}\) denotes forward-pass rate, the effective action rate approaches
\[
f_{\text{eff}} \approx \Kchunk \cdot f_{\text{fwd}}.
\]
The per-time compute budget becomes
\[
\Cbudget \approx \frac{1}{N}\mathrm{FLOPs}_{\text{brain\_once}} + f_{\text{fwd}}\cdot \mathrm{FLOPs}_{\text{cere\_per\_fwd}},
\]
where \(\mathrm{FLOPs}_{\text{cere\_per\_fwd}}\) accounts for processing the \((\Kchunk\times \Dact)\) queries in one pass. We report both \(f_{\text{fwd}}\) and \(f_{\text{eff}}\) for clarity.

\paragraph{Tensor shapes and cached contents.}
Unless stated otherwise, we use the following notation and typical sizes (examples only; exact hyperparameters in Appendix).
\begin{itemize}
  \item Frozen Cerebrum multi-layer outputs: \(H_B^{(l)} \in \mathbb{R}^{T_B^{(l)} \times \dmodel^{(l)}}\), for \(l\in\{l_1,l_m,l_L\}\). Cached in Stage~A.
  \item Layer-wise projections (adapter): \(\tilde{H}^{(l)} = H_B^{(l)} W_l + b_l \in \mathbb{R}^{T_B^{(l)} \times \dmodel}\). Fusion yields \(G \in \mathbb{R}^{T \times \dmodel}\) with \(T=\sum_l T_B^{(l)}\).
  \item Attention pooling (adapter):\\
  \(
  \begin{aligned}
  &Q\in\mathbb{R}^{\Nc\times\dmodel},\quad K_a=GW_k,\quad V_a=GW_v,\\
  &A=\mathrm{softmax}\!\big(QK_a^\top/\sqrt{\dmodel}\big)\in\mathbb{R}^{\Nc\times T},\\
  &\CB=\mathrm{LN}(A V_a)\in\mathbb{R}^{\Nc\times\dmodel}.
  \end{aligned}
  \)
  \item ViT tokens: \(\Vt\in\mathbb{R}^{T_V\times\dmodel}\) (e.g., \(256/16\Rightarrow T_V=16^2=256\) for the main view; each ROI view contributes the same token count). Text tokens: \(\Wt\in\mathbb{R}^{T_W\times\dmodel}\). State token: \(\st\in\mathbb{R}^{1\times\dmodel}\).
  \item Action queries (parameters, not cached): \(\Kchunk\times \Dact\) learnable embeddings \(\{\qk_{k,j}\}\), each produces \(\mathbf{z}_{k,j}\in\mathbb{R}^{\dmodel}\).
  \item Output logits: \(\mathbf{o}_{k,j}\in\mathbb{R}^{3}\) reshaped to \((\text{batch},\Kchunk,\Dact,3)\); labels share the same \((\Kchunk,\Dact)\) grid.
  \item Typical defaults: \(\Nc=24\), \(\dmodel{=}1024\), \(\Kchunk{=}20\), \(\Dact{=}16\); we report exact settings per experiment.
\end{itemize}

% (Removed obsolete placeholder figure)

\section{Data and Training}
We describe datasets, labeling, prompts, caching, and training details aligned with our tripartite design. Defaults are provided and tuned per experiment; we report exact settings alongside results.

\paragraph{Data Sources and Splits.}
We combine (i) public robot demonstrations (e.g., LIBERO subsets~\cite{liu2023libero}), (ii) a small set of real desktop manipulation tasks (pick-place, tool-use), and (iii) optional simulation for long-tail coverage. This design follows recent trends in large-scale data-driven robotics, where diverse multi-task, multi-domain datasets and data-reuse frameworks such as BridgeData, DROID, and reward-sketching based offline RL have been shown to improve generalization and data efficiency~\cite{ebert2021bridgedata,khazatsky2024droid,cabi2019sketching}. Unlike recent web-scale efforts that rely on massive cross-embodiment corpora~\cite{openxembodiment2023,nasiriany2024robocasa,grauman2022ego4d}, we deliberately target a more modest data regime that better matches limited-compute labs. Splits are stratified by \textbf{task}; identities and near-duplicates are de-duplicated across train/val/test. Privacy-sensitive segments are filtered. We fix seeds, log environment versions, and ensure deterministic dataloader order. \textbf{Hardware setup:} a dual-arm system with two 7-DoF arms; each arm mounts a dexterous hand used as a gripper with one open/close DoF. Demonstrations are collected via VR teleoperation; the VR pose stream is mapped to joint-space trajectories with rate and safety limits logged alongside data.

\paragraph{Action Labeling (Categorical Deltas).}
From time-stamped poses/commands, we compute per-dimension deltas and quantize signs \(\{-1,0,+1\}\) with deadbands:
\[
\Delta p \text{ (mm) },\ \Delta \theta \text{ (deg)} \Rightarrow
y_{t,j} \in \{-1,0,+1\}.
\]
Our control dimension is \(\Dact{=}16\) (two 7-DoF arms + two gripper open/close DoFs). We use step grids \(\delta_p{=}5\,\mathrm{mm}\) and \(\delta_\theta{=}1^\circ\) with a zero band calibrated from small-motion noise. Outliers (speed spikes, slips) are down-weighted via per-frame weights \(w_t\) from robust statistics; gripper open/close channels adopt symmetric thresholds matched to the angular grid. For micro-horizon reuse, labels are aligned to the \((\Kchunk\times\Dact)\) grid; we optionally apply mild temporal smoothing before quantization to reduce flicker.

\paragraph{Precision-Control Tasks.}
Beyond semantic goals, we introduce \emph{quantitative} objectives (e.g., ``move the object left by 10\,cm'') to encourage measurable spatial understanding and fine-grained control. We define position/orientation error metrics and success thresholds accordingly, and report error distributions alongside success/jitter/jerk.

\paragraph{Structured Prompts for the Frozen Cerebrum.}
We use three prompt templates (concise, extended, JSON). Unless otherwise noted, training and inference both use the JSON template with fields \texttt{goal}, \texttt{constraints}, \texttt{objects}, \texttt{failure\_cases}, \texttt{environment}. During training, we randomize field order with 50\% probability and apply light text normalization (lowercasing, whitespace trimming) and instruction augmentation (synonym substitution) without changing semantics, to improve robustness and reduce template shift.

\paragraph{Feature Caching (Stage A, offline).}
With the Cerebrum fully frozen, we run it offline to extract early/mid/late hidden states \(\{H_B^{(l)}\}_{l\in\{l_1,l_m,l_L\}}\). We persist per-trajectory archives (\texttt{npz/mmap}, float16 by default) containing:
(i) \texttt{version\_hash}, \texttt{tokenizer\_id}, \texttt{prompt\_id}, \texttt{calib\_id};
(ii) the multi-layer tensors with shapes and dtypes; 
(iii) prompt metadata and raw instruction; 
(iv) camera intrinsics and ROI projections; 
(v) trajectory timestamps and categorical labels; 
(vi) checksums and dependency manifests.
We provide a cache validation script (shape/dtype/hash/timestamp checks) and a small viewer for random samples.

\paragraph{Training (Stage B and optional C).}
\emph{Stage B} jointly trains the \Pons{} and the Cerebellum on cached \(\{H_B^{(l)}\}\) plus current frames. Batches are balanced by \textbf{task difficulty}; we use class-weighted cross-entropy with label smoothing and entropy regularization. Image augmentations include random resize-crop, color jitter, Gaussian noise, and motion blur; ROI is projected from calibration and cropped as \textbf{two views} (left/right end-effector) when available, with confidence-aware fallback to the main view. Text encoder is frozen or lightly adapted via a small adapter. 
\emph{Stage C (optional)} lightly tunes bridging layers (e.g., adapter projections) under a reduced learning rate; the Cerebrum remains frozen throughout.
The modular split naturally enables \emph{cerebellum-only RL} in simulation—fine-tune \headname{} while freezing the Cerebrum and the \Pons{}, targeting smoothness and precision improvements without touching high-level semantics.

\paragraph{Execution Policy and Scheduling (Fixed Ratio).}
We adopt \emph{execute-and-reuse} micro-horizons: one forward produces \(\Kchunk\) steps that are executed sequentially without re-forwarding; no early re-planning is used. Unless otherwise stated, we set \(\Kchunk{=}20\). The Cerebrum is invoked once every \(N\) cerebellum chunks (default \(N{=}5\)). Let \(f_{\text{fwd}}\) be the cerebellum forward rate; the effective action rate is \(f_{\text{eff}}\!\approx\!\Kchunk\cdot f_{\text{fwd}}\). This fixed-ratio schedule matches the Method section and amortizes Cerebrum cost.

\paragraph{Optimization and Regularization.}
Unless stated otherwise, we use AdamW (lr \(1\times10^{-4}\), weight decay \(0.05\)), cosine decay with 2k warmup steps, gradient clipping (1.0), label smoothing \(\epsilon{=}0.05\), class weights from inverse-frequency clipped to \([0.5,2.0]\), entropy regularization \(\lambda_H\in[1\text{e-}3,5\text{e-}3]\), temporal KL \(\lambda_T\in[1\text{e-}3,5\text{e-}3]\). Batch size and training steps are set per experiment (e.g., the preliminary LIBERO setting in Section~5 uses 20k steps with batch \(80\times8\) GPUs). Temperature \(\tau\) is annealed (e.g., \(1.5\to0.7\)); hysteresis thresholds \((\theta_{\uparrow},\theta_{\downarrow})\) default to \((0.2,0.2)\); EMA coefficient \(\alpha{=}0.8\). We log per-dimension confusion and jitter statistics.

\paragraph{Hyperparameters and Defaults.}
We summarize key knobs that are varied in ablations (defaults in parentheses):
\begin{itemize}[leftmargin=1em]
  \item \textbf{Cerebrum context}: layers \(\{l_1,l_m,l_L\}\), context tokens \(\Nc\) (24).
  \item \textbf{Cerebellum}: Transformer depth \(L\) (6), \(\dmodel{=}1024\), heads (8), \(\Kchunk{=}20\), \(\Dact{=}16\).
  \item \textbf{Scheduling}: Cerebrum interval \(N\) (5), no early triggers.
  \item \textbf{Vision}: main view \(1028{\times}800\!\to\!256{\times}256\), patch size 16; two ROI crops \(256{\times}256\) (if available).
  \item \textbf{Text}: frozen vs light adapter (default: frozen).
  \item \textbf{Control grid}: \(\delta_p{=}5\,\mathrm{mm}\), \(\delta_\theta{=}1^\circ\), deadband from noise calibration.
  \item \textbf{Stability}: \(\tau\) anneal (1.5→0.7), \(\theta_{\uparrow/\downarrow}\) (0.2), EMA \(\alpha\) (0.8).
\end{itemize}

\begin{table}[t]
  \centering
  \caption{Key hyperparameters (defaults; tuned in ablations).}
  \label{tab:dt-hypers}
  \begin{adjustbox}{max width=\linewidth}
  \begin{tabular}{ll}
    \toprule
    Component & Setting (default) \\
    \midrule
    Cerebrum Layers & \(\{l_1,l_m,l_L\}\), \(\Nc{=}24\) \\
    ViT & Main \(1028{\times}800\!\to\!256^2\), patch 16; two ROI \(256^2\) \\
    Text Encoder & Frozen (adapter optional) \\
    Cerebellum & \(L{=}6, \dmodel{=}1024, \#heads{=}8, \Kchunk{=}20, \Dact{=}16\) \\
    Scheduling & Fixed \(N{=}5\) (no early re-plan) \\
    Control Grid & \(\delta_p{=}5\,\mathrm{mm},\ \delta_\theta{=}1^\circ\) \\
    Stability & \(\tau\!:\!1.5\!\to\!0.7,\ \theta_{\uparrow/\downarrow}{=}0.2,\ \alpha{=}0.8\) \\
    Optimization & AdamW, lr \(1\text{e-}4\), wd 0.05, warmup 2k, clip 1.0 \\
    \bottomrule
  \end{tabular}
  \end{adjustbox}
\end{table}

\paragraph{Reproducibility and Logging.}
We fix seeds, enable deterministic ops when feasible, and log: \texttt{version\_hash}, data split manifests, cache checksums, hyperparameters, VR teleoperation metadata (rate, safety limits), and wall-clock latency. We report compute-normalized success \(\SRcn\), cold Cerebrum latency, per-step Cerebellum latency (incl.\ ROI), and achieved closed-loop frequency \(f_{\text{eff}}\).

\section{Experiments}
We evaluate the tripartite architecture and report preliminary evidence on LIBERO under the official evaluation environment.

\paragraph{Benchmarks and Tasks.}
LIBERO subsets~\cite{liu2023libero}: \textbf{LIBERO-Spatial}, \textbf{LIBERO-Object}, \textbf{LIBERO-Goal}, and \textbf{LIBERO-Long}. Our evaluation focuses on a controlled, mid-scale regime to study compute-aware trade-offs, complementary to large-scale data-driven setups such as BridgeData and DROID or reward-sketching based frameworks~\cite{ebert2021bridgedata,khazatsky2024droid,cabi2019sketching}. 
Real bimanual tasks (planned): \textbf{folding clothes} and \textbf{put X into pot}. 
Precision task (planned): \textbf{move an object by a fixed distance}. 
LIBERO subsets use \textbf{10 tasks each} and \textbf{500 episodes per subset}. Real/precision tasks episode plans: \textbf{folding clothes} (5 tasks, 200 episodes), \textbf{put X into pot} (10 tasks, 400 episodes), and \textbf{move by fixed distance} (10 tasks, 200 episodes).

\paragraph{Evaluation Protocol.}
We follow the official LIBERO evaluation protocol and run the full test suites. For each trained checkpoint, we run evaluation \textbf{5 times} (randomized evaluation runs) and report the averaged success rate. Success criteria and maximum horizon are fixed per task (details in Appendix). Unless otherwise stated, we use \(\Dact{=}16\).

\paragraph{Training Budget.}
For the preliminary LIBERO results reported here, we train for \textbf{20k steps} with batch size \(80\) on each of \textbf{8 GPUs}. Unless stated otherwise, the LIBERO experiments do not enable ROI due to environment integration constraints.

\paragraph{Metrics.}
We report: success rate; jitter rate (per-dimension sign flips per step); jerk; closed-loop frequencies \(f_{\text{fwd}}\) and \(f_{\text{eff}}\); latency (Cerebrum once-call vs Cerebellum per-forward); intervention rate; training wall-clock/throughput (note: Stage-A is reported separately and amortized); and compute-normalized success \(\SRcn\). For head efficiency, we compare \headname{} against diffusion/flowmatching under identical GPU/resolution/batch.

\paragraph{Main Comparisons.}
(1) Single-system baseline (monolithic VLA) vs \textbf{Tripartite (ours, fixed \(N\), \(\Kchunk{=}20\))}. Monolithic VLAs follow designs similar to GR00T and OpenVLA families~\cite{nvidia2025gr00tn1,kim2024openvla}.\\
(2) Action head: continuous regression / diffusion vs \textbf{\headname{} (Parallel Categorical Action Transformer) head}; we provide a timing protocol in the Appendix under matched conditions (GPU/resolution/batch) to verify anticipated efficiency gains for \headname{}, comparing against diffusion/flow-based VLAs~\cite{chi2023diffusionpolicy,black2024pi0,liu2025rdt1b} and OFT-style parallel decoding~\cite{kim2025oft}.\\
(3) Scale study: Cerebrum size (4B/8B/32B) and Cerebellum size (vary \(L,\dmodel,\) heads) under matched steps.

\paragraph{Ablations.}
A1: last-only vs 1/2/3-layer fusion; \(\Nc\in\{8,16,24,32\}\).\\
A2: with/without Cerebellum image/text pathways; two-ROI on/off; main resolution (256 vs 224).\\
A3: stability controls (temperature anneal, entropy cap, hysteresis, EMA); \(\Kchunk\in\{5,10,20\}\).\\
A4: two-stage caching vs end-to-end training (report wall-clock and seed variance).\\
A5: fixed-interval scheduling only; vary \(N\in\{3,5,10\}\) to chart compute–success trade-offs.
\\
\emph{Pons \& RL (prospective).} Pons factorization: structured vs non-structured recoding; early/mid/late layer selection for \(\CB\). Cerebellum-only RL in simulation: freeze the Cerebrum and the \Pons{}, fine-tune \headname{} with rewards targeting smoothness, precision, and latency.

\begin{table}[t]
  \centering
  \caption{Benchmarks, tasks, and episodes.}
  \label{tab:task-summary}
  \begin{adjustbox}{max width=\linewidth}
  \begin{tabular}{lcc}
    \toprule
    Benchmark/Task & Tasks & Episodes \\
    \midrule
    LIBERO-Spatial & 10 & 500 \\
    LIBERO-Object  & 10 & 500 \\
    LIBERO-Goal    & 10 & 500 \\
    LIBERO-Long    & 10 & 500 \\
    Folding clothes & 5 & 200 \\
    Put X into pot  & 10 & 400 \\
    Move by fixed distance & 10 & 200 \\
    \bottomrule
  \end{tabular}
  \end{adjustbox}
\end{table}

\begin{table}[t]
  \centering
  \caption{Training budget and default setup (main results).}
  \label{tab:train-budget}
  \begin{adjustbox}{max width=\linewidth}
  \begin{tabular}{ll}
    \toprule Item & Setting \\
    \midrule
    Eval runs & 5 per checkpoint (averaged) \\
    Steps & 20k, batch \(80\times8\) GPUs \\
    Cerebrum & N1.5 backbone (default); layer is mid-layer (e.g., 12 for Eagle2.5) \\
    Head & N1.5 FlowMatching head (kept fixed unless stated) \\
    Scheduling & \(N{=}1\) for LIBERO comparisons; \(\Kchunk{=}16\) for SaiVLA0 \\
    Vision & official LIBERO setup (ROI not enabled in LIBERO) \\
    \bottomrule
  \end{tabular}
  \end{adjustbox}
\end{table}

\ifshowresults
\paragraph{Preliminary LIBERO Results.}
We report success rates (\%) on the official LIBERO evaluation environment (full subsets), averaged over 5 evaluation runs per checkpoint. These results are intended as stage-wise evidence; ROI is used in real-robot experiments but is not enabled in LIBERO in this draft.

\begin{table}[t]
  \centering
  \small
  \caption{Split (feature caching) vs official N1.5 head-only training on LIBERO. Split caches the mid-layer hidden states per frame (Stage A, offline; \(\sim\)1h one-time for LIBERO) and trains the downstream head without backbone forward (Stage B). Training time is reported under the same hardware setup. Goal shows a small drop while Long improves substantially.}
  \label{tab:libero-split}
  \begin{adjustbox}{max width=\linewidth}
  \begin{tabular}{lcccccc}
    \toprule
    Method & Spatial & Object & Goal & Long & Mean & Train (h) \\
    \midrule
    GR00T-N1.5 (official) & 92.0 & 92.0 & 86.0 & 76.0 & 86.5 & 7.5 \\
    GR00T-N1.5 (split)    & 97.8 & 99.6 & 79.6 & 92.8 & 92.5 & 4.5 \\
    \bottomrule
  \end{tabular}
  \end{adjustbox}
\end{table}

\begin{table}[t]
  \centering
  \small
  \caption{Backbone comparison under the same split training and the same N1.5 FlowMatching head. Eagle2.5 uses layer 12 and Qwen3VL-2B uses layer 14 (both mid-layers).}
  \label{tab:libero-backbone}
  \begin{adjustbox}{max width=\linewidth}
  \begin{tabular}{lccccc}
    \toprule
    Backbone & Spatial & Object & Goal & Long & Mean \\
    \midrule
    Eagle2.5VLM (FM0) & 97.8 & 99.6 & 79.6 & 92.8 & 92.5 \\
    Qwen3VL-2B (FM0)  & 95.8 & 94.8 & 82.0 & 87.6 & 90.1 \\
    \bottomrule
  \end{tabular}
  \end{adjustbox}
\end{table}

\begin{table}[t]
  \centering
  \small
  \caption{LIBERO success rates (\%) for SaiVLA0 and several VLA models. SaiVLA0 uses Tripartite + \headname{} (ParaCAT), with \(N{=}1\), \(\Kchunk{=}16\), and no ROI in LIBERO. Note: All data in the table are sourced from original papers, official project homepages, or academic papers.}
  \label{tab:libero-compare}
  \begin{adjustbox}{max width=\linewidth}
  \begin{tabular}{lccccc}
    \toprule
    Method & Spatial & Object & Goal & Long & Mean \\
    \midrule
    SaiVLA0 (ours) & 99.8 & 100.0 & 98.2 & 97.8 & 99.0 \\
    $\pi_0$~\cite{black2024pi0} & 96.8 & 98.8 & 95.8 & 85.2 & 94.2 \\
    OpenVLA-OFT~\cite{kim2025oft} & 97.6 & 98.4 & 97.9 & 94.5 & 97.1 \\
    GR00T-N1.6~\cite{nvidia2025gr00tn1_5} & 97.7 & 98.5 & 97.5 & 94.4 & 97.0 \\
    $\pi_0.5$ & 98.8 & 98.2 & 98.0 & 92.4 & 96.9 \\
    GR00T-N1.5 (official) & 92.0 & 92.0 & 86.0 & 76.0 & 86.5 \\
    \bottomrule
  \end{tabular}
  \end{adjustbox}
\end{table}

\begin{figure}[t]
  \centering
  \IfFileExists{figures/Real_Robot.png}{
    \includegraphics[width=\linewidth]{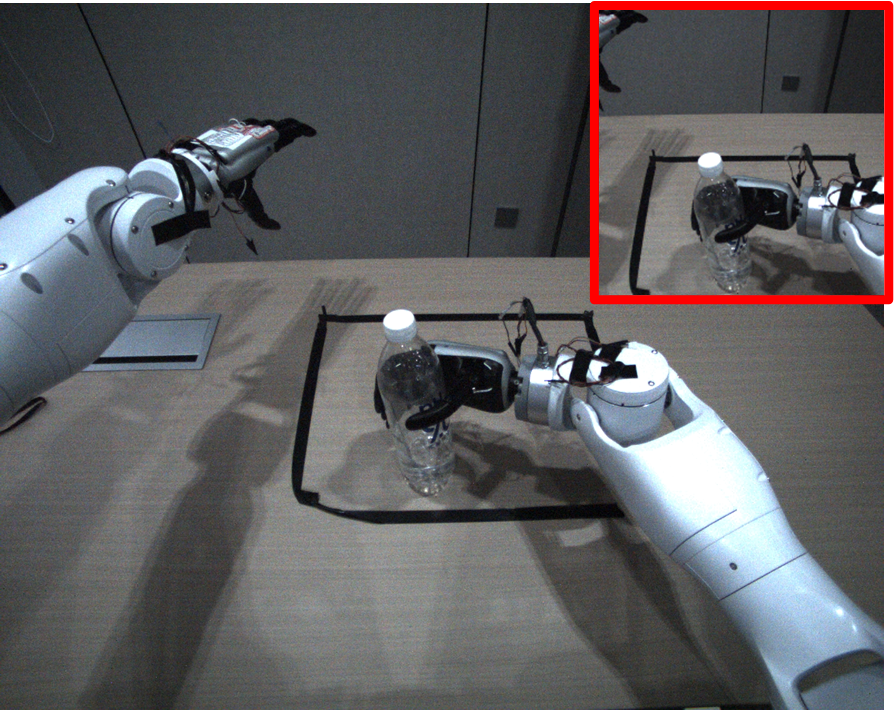}
  }{
    \fbox{\parbox[c][0.22\textheight][c]{0.95\linewidth}{\centering Real-robot snapshot placeholder\\\texttt{figures/Real\_Robot.png}}}
  }
  \caption{Real-robot simple grasping: grasping a white bottle. The top-right inset visualizes the geometry-tied ROI view used on real robot; LIBERO experiments in this draft do not enable ROI.}
  \label{fig:real-bottle}
\end{figure}
\fi

\section{Limitations \& Ethics}
We list limitations, ethical considerations, and threats to validity aligned with our tripartite setup.

\paragraph{Limitations.}
(i) \textbf{Frozen Cerebrum domain shift.} The 8B VLM is fully frozen; strong visual/task shift or prompt mismatch can reduce the utility of \(\CB\) tokens.\\
(ii) \textbf{No early re-planning under fixed scheduling.} We adopt fixed-interval Cerebrum calls (\(N{=}5\)) and micro-horizon reuse (\(\Kchunk{=}20\)); the lack of uncertainty-based re-planning may hurt adaptivity under rapid changes.\\
(iii) \textbf{Categorical precision ceiling.} Per-dimension \(\{-1,0,+1\}\) deltas can bottleneck sub-millimeter/degree docking; a hybrid head (classification + residual regression) may be required.\\
(iv) \textbf{Class imbalance and calibration.} The 0-class often dominates; despite class weights and label smoothing, careful temperature/hysteresis calibration remains important.\\
(v) \textbf{Dual-arm coordination.} Bimanual coupling (7+7 DoF plus 2 grippers, \(\Dact{=}16\)) increases contact/constraint complexity; small label noise can cause asymmetric oscillations.\\
(vi) \textbf{ROI/calibration sensitivity.} Wrist ROIs (two \(256^2\) crops) depend on accurate intrinsics/extrinsics and time sync; drift or occlusion degrades ROI quality (we fallback to the main view but may lose fine contact cues).\\
(vii) \textbf{Cache consistency.} Two-stage training relies on frozen Cerebrum caches; any change to tokenizer/prompt/calibration/layer picks invalidates caches (we enforce version hashes and validators, yet accidental staleness is a risk).\\
(viii) \textbf{Compute–latency trade-offs.} Larger \(\Nc,L,\dmodel\) or smaller \(N\) can improve success but increase latency/compute; \(\Kchunk\) affects \(f_{\text{eff}}\) vs reactivity. Our compute-normalized score (\(\SRcn\)) mitigates but does not eliminate these trade-offs.\\
(ix) \textbf{Evaluation averaging.} We average results over 5 evaluation runs per checkpoint and do not report mean/std explicitly; while averaging reduces noise, it may hide dispersion and under-report instability.
(x) \textbf{Split training is not strictly engineering-equivalent in practice.} While split feature caching (offline hidden-state caching + downstream head training) is intended as an engineering optimization under frozen-backbone, head-only training, we observe non-trivial success differences on LIBERO (including a small drop on Goal). The reason is currently unclear and may relate to implementation, numerical details, or data/IO pipelines; a more exhaustive audit is left for future work.\\

\paragraph{Ethics and Safety.}
(1) \textbf{Bimanual safety.} We enforce torque/joint/velocity limits, workspace guarding, collision checks, e-stop, and human-in-the-loop supervision during development/real trials. Dual grippers introduce pinch hazards; we follow standardized risk assessments before execution.\\
(2) \textbf{Data privacy and consent.} VR teleoperation logs may contain operator-related signals; we collect with informed consent, avoid face/audio capture, and strip any PII. Public datasets (e.g., LIBERO) are used under their licenses; real-task data follows site policies.\\
(3) \textbf{Release policy.} We plan to release configs, prompt templates, cache schemas, and validators. Precomputed caches may be released after redaction and license checks; raw videos/logs with potential PII will not be released.\\
(4) \textbf{Misuse prevention.} This system targets research on low-latency control under small data. It is not a safety-certified controller; deployment in uncontrolled environments is out of scope. We document intended use and known risks.\\
(5) \textbf{Environmental impact.} We report compute and latency metrics to promote compute-aware comparisons; scaling studies (4B/32B) are run under matched steps.

\paragraph{Threats to validity.}
(a) \textbf{Benchmark coverage.} Results depend on LIBERO-Spatial/Object/Goal/Long (each 10 tasks, 500 episodes) and three real/precision suites (folding clothes: 5/200; put X into pot: 10/400; move by fixed distance: 10/200). Findings may not generalize to other robots/sensors.\\
(b) \textbf{Calibration and ROI.} Small calibration drift or timestamp skew can bias attention to ROIs; our fallback mitigates but does not remove bias.\\
(c) \textbf{Metric definitions.} Jitter/jerk definitions and success criteria can influence conclusions; we pre-define protocols and release scripts.\\
(d) \textbf{Budget matching.} We match training steps across methods; different schedulers/heads may benefit from different budgets. We therefore also report \(\SRcn\), \(f_{\text{fwd}}\), \(f_{\text{eff}}\), and latencies, yet residual bias may remain.\\
(e) \textbf{Prompt/template effects.} Structured JSON prompts reduce variance but template choice still impacts \(\CB\); we randomize field order and document templates.\\
(f) \textbf{Hyperparameter sensitivity.} Performance can vary with \(\Nc,L,\dmodel,\text{heads},\Kchunk,N\); we provide grids and ablations, but full coverage is infeasible.

\section{Conclusion \& Future Work}
We introduced SaiVLA-0, a tripartite VLA framework that freezes a large VLM (Cerebrum) for low-frequency semantic planning and equips a multi-modal Cerebellum with categorical control for high-frequency execution. Our design uses fixed-ratio scheduling (default Cerebrum cadence \(N{=}5\)) and micro-horizon reuse (default \(\Kchunk{=}20\)), a trainable Pons Adapter, two-stage feature caching for reproducibility, and dual-view ROI to improve contact cues. In our stage-wise evaluation (Section~5), we report preliminary evidence on LIBERO and average over 5 evaluation runs per checkpoint. This separation of concerns reduces engineering overhead during iteration for limited-data regimes and makes the latency–stability–compute trade-offs explicit.

\paragraph{Future Work.}
(i) \textbf{Adaptive scheduling and re-planning.} Move beyond fixed \(N\) to uncertainty/failure-triggered or learned policies (RL/active learning), with calibrated confidence and early re-plan under distribution shift~\cite{barto2003hrl,sutton2018reinforcement}.\\
(ii) \textbf{Hybrid action heads and adaptive grids.} Combine categorical deltas with residual regression for sub‑mm/deg precision; explore adaptive step sizes, dynamic \(\Kchunk\), and mixed discrete–continuous decoding~\cite{chi2023diffusionpolicy,black2024pi0,pertsch2025fast}.\\
(iii) \textbf{ROI robustness and online calibration.} Multi-view fusion, self-checks for intrinsics/extrinsics/time sync, and automatic ROI fallback/repair to mitigate drift/occlusion.\\
(iv) \textbf{Scaling and compression.} Systematic scaling laws across Cerebrum sizes (4B/8B/32B) and Cerebellum capacity (\(\Nc,L,\dmodel,\) heads) under compute-normalized metrics; edge deployment via pruning/quantization/distillation.\\
(v) \textbf{Data and supervision.} Expand VR teleoperation to semi/self‑supervised pipelines, improve label denoising for bimanual coordination, and leverage simulation for task coverage with sim2real regularization.\\
(vi) \textbf{Safety and guarantees.} Integrate constraint-aware decoding and certified safety monitors for bimanual manipulation; document deployment guidelines beyond lab settings.\\
(vii) \textbf{Generalization.} Evaluate on broader robot platforms and task suites (incl. long-horizon bimanual), with cross-domain prompts and structured template search.\\
(viii) \textbf{Open tooling.} Harden cache schema, validators, and config recipes for plug-and-play reproducibility and fair compute-aware comparison.

\balance
\small
\bibliographystyle{unsrtnat}
\bibliography{references}

\appendix
\section*{Appendix}
% Appendix: structured prompts, cache schema, evaluation definitions (placeholders)

\paragraph{Structured Prompt Templates.}
We provide three templates: \emph{concise}, \emph{extended}, and \emph{JSON}. Unless otherwise stated, training and inference use the JSON template. During training, JSON field order is randomized with probability 0.5 to improve robustness; inference matches training templates to reduce shift.\\
\textbf{Concise template (one-line; used in ablations or quick setups):} \texttt{what the robot do to :\{task\}}.\\
\textbf{JSON fields (when structured prompts are used):} \texttt{goal}, \texttt{constraints}, \texttt{objects}, \texttt{failure\_cases}, \texttt{environment}.

\paragraph{Cache Schema (Cerebrum Stage A).}
Stored as \texttt{npz/mmap} with fields: \texttt{version\_hash}, \texttt{dataset\_id}, \texttt{task\_id}, \texttt{big\_brain\_id}, \texttt{tokenizer\_id}, \texttt{prompt\_id}, \texttt{prompt\_hash}, \texttt{layers}, \(\Nc\), \(\dmodel\), \texttt{C} (path), \texttt{traj} (path), \texttt{K\_chunk}, \texttt{camera\_calib} (K,R,t), \texttt{roi\_meta}, \texttt{timestamp}, \texttt{checksum}. Validation checks version equality, shape consistency, hash matches, and NaN absence.

\paragraph{Evaluation Protocols.}
We define success rate, jitter rate (sign flips per step), jerk, intervention rate, and report latency split for Cerebrum vs Cerebellum. Compute-normalized success is \(\SRcn = \mathrm{SuccessRate}/\Cbudget\), where \(\Cbudget = \frac{1}{N}\mathrm{FLOPs}_{\text{brain\_once}} + f \cdot \mathrm{FLOPs}_{\text{cere\_per\_step}}\) with \(f\) the control frequency and \(N\) the Cerebrum interval.

\paragraph{Timing Protocol (Inference \& Training).}
Inference: we fix GPU, resolution, and batch; measure single-step control latency for \headname{} (single forward) and diffusion/flowmatching heads (multi-step sampling), repeating runs and reporting median. Training: we time Stage~B steps/sec under matched steps/batch; Stage~A cache generation is reported separately and amortized for fair iteration-throughput comparison. As a concept-and-protocol paper with preliminary evidence, we provide this timing protocol to verify anticipated efficiency gains under matched conditions.

\paragraph{Pons details and factorization criteria.}
We instantiate the \Pons{} as a semantic-to-dynamics compiler that (i) sparsifies and recodes structured Cerebrum intent into execution-ready tokens, (ii) factorizes action structure into geometry, dynamics priors, and control objectives to form composable motor primitives, and (iii) aligns feedback and intent to ease cerebellar forward-model updates. We document token shapes and layer picks, and provide an ablation grid on factorization choices.

\paragraph{Cerebellum-only RL protocol (prospective).}
We freeze the Cerebrum and the \Pons{}, and fine-tune \headname{} in simulation with rewards combining success, smoothness (jerk/jitter regularizers), precision error terms, and mild latency penalties. Safety constraints (velocity/acceleration bounds) are enforced. This protocol tests whether RL can further refine low-latency control without touching high-level semantics.

\paragraph{Main default hyperparameters (this work).}
Layers \(\{l_1,l_m,l_L\}\); \(\Nc{=}24\); \(\dmodel{=}1024\). Vision: ViT‑S/16, main \(256{\times}256\) (from \(1028{\times}800\)), two ROI \(256{\times}256\). Text encoder frozen (adapter optional). Cerebellum: \(L{=}6\), \(\dmodel{=}1024\), heads\(=8\), \(\Kchunk{=}20\); \(\Dact{=}16\). Control grid: \(\delta_p{=}5\,\mathrm{mm}\), \(\delta_\theta{=}1^\circ\). Stability: \(\theta_{\uparrow/\downarrow}{=}0.2\), EMA \(\alpha{=}0.8\), temperature annealing \(\tau: 1.5\to 0.7\), entropy cap \(\Hmax{=}0.9\). Scheduling: fixed \(N{=}5\). These match Table~\ref{tab:dt-hypers} and Section~5.

\paragraph{Historical prototype settings (not used in main results).}
Early single‑arm prototypes explored smaller grids and resolutions (e.g., main 224, ROI 128; \(K{=}3\); \(\Dact{=}7\); \(\delta_{pos}{\approx}1\,\mathrm{cm}\), \(\delta_{rot}{\approx}2^\circ\)). We omit those numbers from main comparisons; ablations are reported under the unified main defaults above.

\end{document}